\documentclass[letterpaper,conference]{IEEEtran}
\IEEEoverridecommandlockouts
\usepackage{cite}
\usepackage{amsmath,amssymb,amsfonts}
\usepackage{algorithmic}
\usepackage{graphicx}
\usepackage{textcomp}
\usepackage{xcolor}
\usepackage{array}
\usepackage{dblfloatfix}
\usepackage{subcaption}
\usepackage{multicol}
\usepackage{lipsum}
\usepackage{booktabs}
\usepackage{multirow}
\usepackage{comment}
\usepackage{mathptmx} 
\usepackage{times} 

\def\BibTeX{{\rm B\kern-.05em{\sc i\kern-.025em b}\kern-.08em
    T\kern-.1667em\lower.7ex\hbox{E}\kern-.125emX}}

\begin{document}

\title{On Interpretability of Deep Learning based Skin Lesion Classifiers using Concept Activation Vectors\\
}


\author{\IEEEauthorblockN{Adriano~Lucieri}
\IEEEauthorblockA{\textit{{Hochschule Pforzheim}} and \\
\textit{Smart Data and Knowledge Services} \\
\textit{German Research Center for Artificial}\\
\textit{Intelligence GmbH (DFKI)}\\
Kaiserslautern, Germany \\
{0000-0003-1473-4745}}
\and
\IEEEauthorblockN{Muhammad~Naseer~Bajwa}
\IEEEauthorblockA{\textit{Technische Universit\"at Kaiserslautern} and\\
\textit{Smart Data and Knowledge Services} \\
\textit{German Research Center for Artificial}\\
\textit{Intelligence GmbH (DFKI)}\\
Kaiserslautern, Germany \\
0000-0002-4821-1056}
\and
\IEEEauthorblockN{Stephan~Alexander~Braun}
\IEEEauthorblockA{\textit{Department of Dermatology} \\
\textit{Universitätsklinikum M\"unster} and\\
\textit{Department of Dermatology} \\
\textit{Universitätsklinikum D\"usseldorf}\\
Münster, Germany \\
0000-0002-5600-0195}
\and
\IEEEauthorblockN{Muhammad Imran Malik}
\IEEEauthorblockA{\textit{National University of Science and} \\
\textit{Technology (NUST)} and\\
\textit{National Center of Artificial Intelligence}\\
Islamabad, Pakistan \\
0000-0002-8079-5119}
\and
\IEEEauthorblockN{Andreas Dengel}
\IEEEauthorblockA{\textit{Technische Universit\"at Kaiserslautern} and\\
\textit{Smart Data and Knowledge Services} \\
\textit{German Research Center for Artificial}\\
\textit{Intelligence GmbH (DFKI)}\\
Kaiserslautern, Germany \\
0000-0002-6100-8255}
\and
\IEEEauthorblockN{Sheraz Ahmed}
\IEEEauthorblockA{\textit{Smart Data and Knowledge Services} \\
\textit{German Research Center for Artificial}\\
\textit{Intelligence GmbH (DFKI)}\\
Kaiserslautern, Germany \\
0000-0002-4239-6520}
}

\maketitle

\begin{abstract}
Deep learning based medical image classifiers have shown remarkable prowess in various application areas like ophthalmology, dermatology, pathology, and radiology. However, the acceptance of these Computer-Aided Diagnosis (CAD) systems in real clinical setups is severely limited primarily because their decision-making process remains largely obscure. This work aims at elucidating a deep learning based medical image classifier by verifying that the model learns and utilizes similar disease-related concepts as described and employed by dermatologists. We used a well-trained and high performing neural network developed by REasoning for COmplex Data (RECOD) Lab for classification of three skin tumours, i.e. Melanocytic Naevi, Melanoma and Seborrheic Keratosis and performed a detailed analysis on its latent space. Two well established and publicly available skin disease datasets, PH\textsuperscript{2} and derm7pt, are used for experimentation. 
Human understandable concepts are mapped to RECOD image classification model with the help of Concept Activation Vectors (CAVs), introducing a novel training and significance testing paradigm for CAVs. Our results on an independent evaluation set clearly shows that the classifier learns and encodes human understandable concepts in its latent representation. Additionally, TCAV scores (Testing with CAVs) suggest that the neural network indeed makes use of disease-related concepts in the correct way when making predictions. We anticipate that this work can not only increase confidence of medical practitioners on CAD but also serve as a stepping stone for further development of CAV-based neural network interpretation methods.
\end{abstract}


\let\svthefootnote\thefootnote
\let\thefootnote\relax\footnotetext[1]{This work is funded by National University of Science and Technology (NUST), Pakistan through Prime Minister's Programme for Development of PhDs in Science and Technology, BMBF projects ExplAINN (01IS19074) and DeFuseNN (01IW17002).}
\let\thefootnote\relax\footnote{\copyright 2020 IEEE. Personal use of this material is permitted. Permission from IEEE must be obtained for all other uses, in any current or future media, including reprinting/republishing this material for advertising or promotional purposes, creating new collective works, for resale or redistribution to servers or lists, or reuse of any copyrighted component of this work in other works.} 
\addtocounter{footnote}{-1}\let\thefootnote\svthefootnote

\section{Introduction}
\label{intro}

United Nations (UN) has recognised healthcare and well-being as one of the 17 Sustainable Development Goals (SDGs) to create a better future for all by 2030 \cite{UN17Goals}. However, achieving this goal requires concerted and sustained efforts in utilizing all available means to improve healthcare since many people are needlessly suffering from preventable diseases. In 2017, AI for Good~\cite{AI4G2017}, a UN initiative to provide a global platform for researchers, identified great potential of Artificial Intelligence (AI) to achieve these SDGs and to help solve the greatest global challenges.

Numerous remarkable studies have been conducted in the last few years successfully applying deep learning for disease classification using various medical image modalities \cite{esteva2017dermatologist,bajwa2019two,litjens2017survey}. However, the acceptance of such Computer-Aided Diagnosis (CAD) solutions with doctors and patients remains dubious at best due to the fact that the process behind learning and encoding features in latent space by computer models is not very well understood. This lack of transparency in the whole decision-making process cannot be overlooked in various critical application areas including medical diagnosis. Especially after Europe's General Data Protection Regulations (GDPR)~\cite{GDPR2018} came into effect in 2018, data subjects are entitled to \emph{Right to Explanation} for any automated decision made by computer algorithms. It is, therefore, need of the hour to elucidate the working principle of deep learning based classifiers so that practical applications of AI in medical diagnosis can be realized expeditiously.

Compared to other fields of applications of Deep Neural Networks (DNNs), medical image analysis often presents unique challenges due to inherent complexity of this task. Manual classification of complex diseases involves recognizing subtle features and high-level concepts that are challenging to grasp without expert knowledge. Even with expert knowledge, doctors' subjective understanding of disease biomarkers leads to low inter-expert agreement ~\cite{fortina2012s, corona1996interobserver}. Therefore, common explanation methods like visualization of saliency maps, which strongly rely on spatial divisibility of concepts, work well on common object detection tasks~\cite{ghorbani2019deep, jolly2018convolutional, selvaraju2016grad} that have well distinguishable features, but fail on more complex medical image analysis tasks.

Skin cancer is the most common type of cancer in the U.S~\cite{ACS2017Facts}. According to a recent study \cite{siegel2019cancer}, skin cancer related death rate forecast for U.S. in 2019 amounted to 11,650 people. These rising rates of skin cancer incidences can not only cost precious lives but also incur huge burden on healthcare systems. It is estimated that approximately 3 million people are treated annually for skin cancer in the U.S. and it costs around 8.1 billion USD~\cite{ruiz2019analysis}.

In this study, we choose classification of skin diseases as a use case to understand what DNNs learn and what they rely on for their predictions in medical diagnosis. We attempt to understand if the \emph{concepts} learnt by classifiers in complex Medical Image Analysis (MIA) tasks are similar to those used by dermatologists. We use two publicly available datasets of dermoscopic images to learn concept mappings i.e. PH\textsuperscript{2} and derm7pt. These datasets are selected because they provide concept annotations in addition to image-wise diagnosis labels. Summarizing our contributions, this study presents;

\begin{itemize}
    \item A new training and significance testing paradigm for Concept Activation Vectors (CAVs) using identically distributed data.
    \item Mapping of concepts learnt by a deep model, in its latent space for skin lesion classification, to dermatologically significant human-understandable concepts using CAVs.
    \item Analysis of contributions of different dermoscopic criteria to the predictions of deep models, revealing agreement between reasoning process of doctors and deep models.
\end{itemize}  

\section{Related Work}
\label{sec:relatedWork}
Understanding the way neural networks learn and explaining their prediction behaviour are active areas of research \cite{palacio2018deep,samek2017explainable}. To interpret these inherently nonlinear mathematical models, three main types of methods are usually employed.

\subsection{Saliency-Based Neural Network Explanations}
Saliency-based methods for neural network explanation were among the first tools towards explainable AI and are still highly in use today. Examples for those methods are GradCAM~\cite{selvaraju2016grad}, SmoothGrad~\cite{smilkov2017smoothgrad}, Integrated Gradient~\cite{sundararajan2017axiomatic} and Layer-Wise Relevance Propagation (LRP)~\cite{bach2015pixel} to name a few. Since these methods create importance maps based on individual input samples, they provide only local interpretations and are unable to explain network's decisions on a global scale. 

To date, there are only very few works focusing on interpreting deep classifiers for dermoscopic images using saliency-based attribution methods ~\cite{young2019deep, wu2018evidence}. This might partly be due to the innate difficulty in skin lesion classification that mandates huge amount of expert knowledge to recognize complex and subtle structures. It could also be due to a large variation in fine nuances of these structures that are hard to discern yet can drastically change diagnosis. Moreover, the visual artefacts corresponding to various diseases in skin lesion images sometimes overlap and are usually distributed all over the image, which does not fare well with saliency-based model interpretation. Other domains in which these methods are frequently used for interpretation like object classification~\cite{ghorbani2019deep, jolly2018convolutional, selvaraju2016grad} usually show more discriminatory features corresponding to specific parts of an object -- for instance tires or headlights of cars.

\subsection{Text-Based Neural Network Explanations} Textual explanation methods for neural networks can be either template-based~\cite{munir2019tsxplain, guo2018neural, anne2018grounding} that generate justifications from some auxiliary information or rule-based~\cite{hancock2018training, srivastava2017joint, zhang2017mdnet, jing2017automatic}, where a classifier is trained with images as well as additional natural language explanations. Zhang et al.~\cite{zhang2017mdnet} proposed a unified network called MDNet following a rule-based approach that generates diagnostic reports along with corresponding attention maps of input images in order to increase the semantic and visual interpretability of MIA task at hand. Jing et al.~\cite{jing2017automatic} proposed a multi-task learning framework that is able to localize abnormal regions in medical images, predict tags and generate their descriptions.

\subsection{Concept-Based Neural Network Explanations} 
Concept-based explanations address the problem of explaining black-box models by finding human-understandable concepts in neural networks' latent representation. Kim et al. \cite{kim2017interpretability} introduced \textit{Concept Activation Vectors} (CAVs) that are used to map human-understandable concepts to latent representation learnt by models in a supervised way using general human concept patches that were taken from various domains. By calculating these main concept directions and by leveraging directional derivatives, they were able to quantify the influence of a concept to the prediction of single output classes.
Zhou et al. \cite{zhou2018interpretable} followed a similar approach by decomposing neural networks' activations into semantically meaningful components pre-trained from a large concept corpus. Graziani  et  al. \cite{graziani2018regression, graziani2019improved} restated the problem of classification in CAVs and employed regression instead. They applied their so-called \textit{Regression Concept Vectors} (RCVs) on problems from medical domain like binary classification of breast cancer histopathology slides and classification of Retinopathy of Prematurity (ROP) states. As an extension to the work in \cite{kim2017interpretability}, Ghorbani  et  al.~\cite{ghorbani2019automating} recently developed a method for unsupervised clustering of object datasets by first applying segmentation of single objects and then clustering activations of object patches into semantically meaningful clusters.

\bigbreak
To the best of our knowledge, concept-based explanation methods have not previously been explored for skin lesion classification networks. Due to the nature of this problem, not all of the previously described methods can be directly applied on this task. Unsupervised clustering as used in~\cite{ghorbani2019automating}, for example, is not suitable in skin lesions as there is a huge spatial concept overlap and thus no possibility for distinct part segmentation. RCVs are also not applicable as skin lesion concepts are hardly quantifiable. The method in~\cite{zhou2018interpretable} requires a concept corpus which is not readily available for this specific task. Any type of textual explanation generation is also not applicable, as no diagnostic reports or descriptions of diagnosis are provided with any public dermoscopic skin lesion dataset. The computation of CAVs as seen in~\cite{kim2017interpretability} requires patches corresponding to general human-understandable concepts. In this work we adopt the TCAV method to the problem of skin lesion classification. Instead of providing general, out-of-distribution concept patches, we train CAVs using samples from identically distributed datasets to map human-understandable concepts to the network's latent space.
\begin{figure*}[h!]%
\centering
\begin{subfigure}[t]{0.195\textwidth}
\includegraphics[width=\columnwidth]{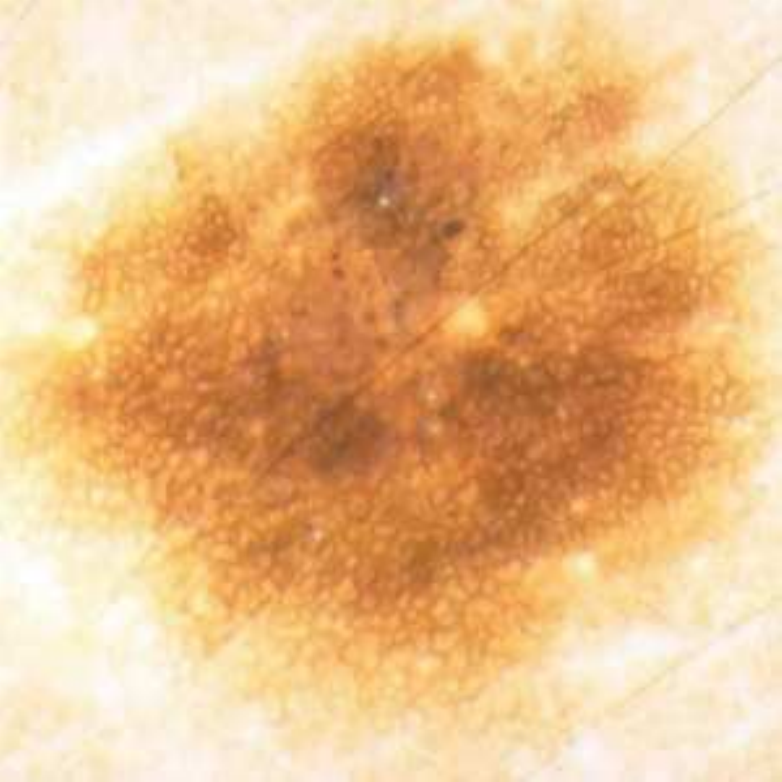}%
\caption{Typical Pigment Network}%
\label{subfig-PNA}%
\end{subfigure}
\begin{subfigure}[t]{0.195\textwidth}
\includegraphics[width=\columnwidth]{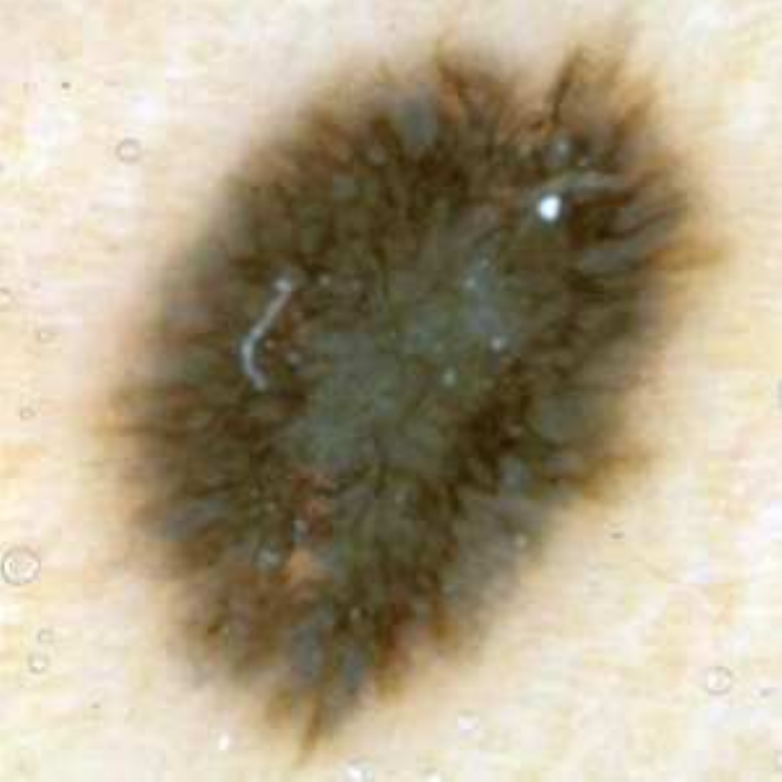}%
\caption{Regular Streaks}%
\label{subfig-SR}%
\end{subfigure}
\begin{subfigure}[t]{0.195\textwidth}
\includegraphics[width=\columnwidth]{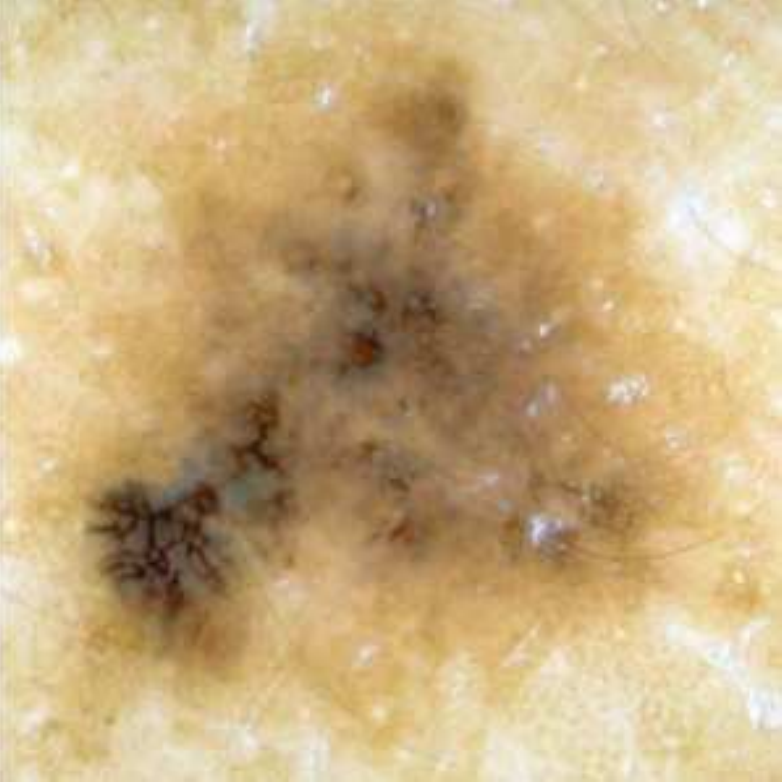}%
\caption{Regression Structure}%
\label{subfig-RS}%
\end{subfigure}
\begin{subfigure}[t]{0.195\textwidth}
\includegraphics[width=\columnwidth]{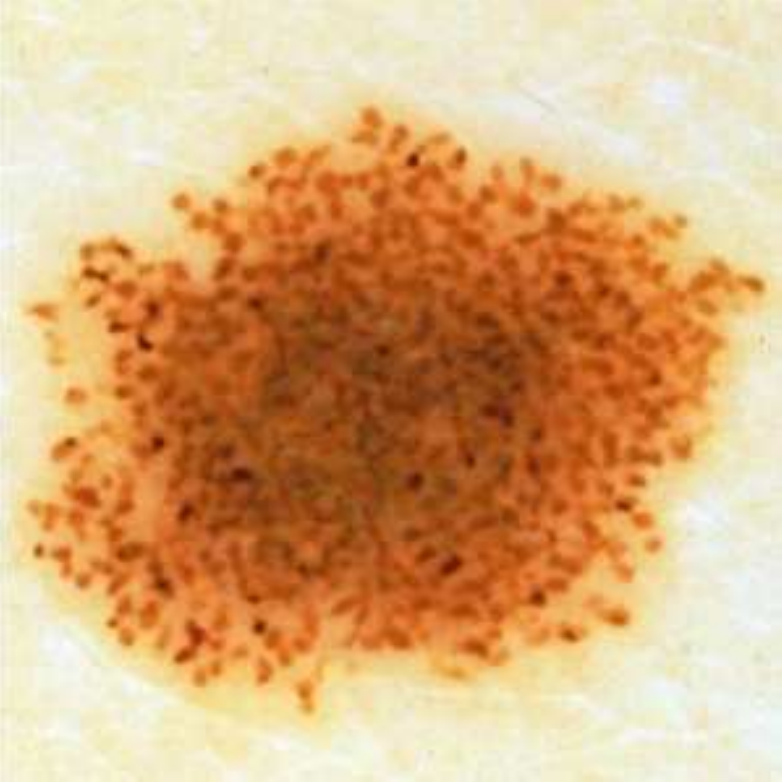}%
\caption{Regular Dots \& Globules}%
\label{subfig-DnG}%
\end{subfigure}%
\begin{subfigure}[t]{0.195\textwidth}
\includegraphics[width=\columnwidth]{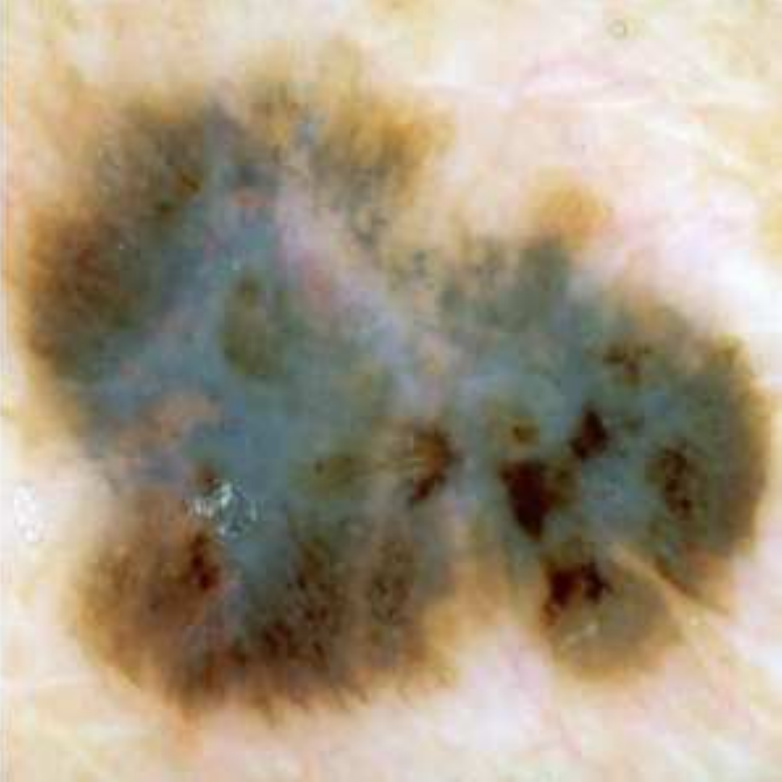}%
\caption{Blue Whitish Veil}%
\label{subfig-BWV}%
\end{subfigure}%
\caption{Exemplary cases of skin lesion concepts from derm7pt dataset.}
\label{fig:conceptExp}
\end{figure*}
\section{Background}
\label{sec:background}

This section briefly describes CAVs and the method of calculating TCAV scores used in this work to quantify the contribution of a concept to DNN's prediction. Moreover, dermoscopic concepts explaining the classifier's decisions are also introduced.

\subsection{Concept Activation Vectors}
\label{subsec:cav}
In order to achieve human-centered interpretability of DNNs, Kim et al.~\cite{kim2017interpretability} introduced \textit{Concept Activation Vectors}. A CAV, $\vec{v}_c$, is a vector in the embedding space of a neural network pointing into the direction that encodes the given concept $c$. CAVs can be calculated by training a binary concept classifier dividing samples containing a given concept from samples where the concept is absent. The CAV is then defined as the normal to the hyperplane separating the two classes.

\paragraph{TCAV Score} The metric introduced in~\cite{kim2017interpretability} to estimate the influence of a CAV on a class of input images is the TCAV score. It makes use of directional derivatives $S_{C,k,l}(x)$ to measure the contextual sensitivity of a concept towards an entire input class, therefore providing global explanations. The TCAV score is given by:

\begin{equation}
    TCAV_{Q_{C,k,l}} = \frac{|{x \in X_k : S_{C,k,l}(x) > 0}|}{|X_k|},
\end{equation}

where $X_k$ denotes all inputs, $k$ represents the class labels and $S_{C,k,l}(x)$ the directional derivate of a sample's activation $x$ from layer $l$ with respect to class $k$ and concept $C$. The TCAV score effectively measures the ratio of class $k$'s inputs, that are positively affected by concept $C$ without taking any magnitude into account. As compared to saliency maps or other per-feature metrics, the TCAV score allows for quantitative evaluation of concepts on whole input classes.

\subsection{Dermoscopic Concepts used for Analysis}
\label{subsec:concepts}
The concepts used in this work to interpret the deep classifier are briefly defined below in accordance with standardized terminology agreed upon by expert dermatologist in 3rd Consensus Conference of the International Society of Dermoscopy (IDS)~\cite{kittler2016standardization}. Fig.~\ref{fig:conceptExp} depicts examples of some concepts listed below.

\subsubsection{Pigment Networks} Pigment Networks consist of interconnected pigmented lines forming a gridlike pattern. Depending on the subtype of Pigment Network, it can either have minimal variability in colour, thickness and spacing of the lines, forming a symmetric grid (Typical Pigment Network) or have greater variability in colour, thickness and spacing of the lines, forming an asymmetric grid (Atypical Pigment Network). Apart from those two general types, more subtypes are also defined in literature. Atypical Pigment Networks can be a clue for Melanoma (although many dysplastic naevi also have atypical networks) whereas typical Pigment Networks normally indicate benign melanocytic lesions (Naevi).

\subsubsection{Streaks} Streaks describe an abnormality of the lesion that can either have the form of pure straight radial extensions, radial extensions with bulbous and often kinked projections on their ends, or a widening of broken lines with incomplete connections. Streaks are referred to as irregular if they are irregularly distributed along the edge of the lesion and are brown-black in colour~\cite{argenziano2003dermoscopy}. Regular Streaks indicate benign lesions and Irregular Streaks are clues for malignant Melanoma.
 
\subsubsection{Regression Structures} Regression Structures are characterized by the appearance of either areas of fine, grey-blue dots, or areas of skin whiter than the surrounding normal-looking skin without blood vessels or shiny-white structures. Its presence is highly indicative of melanoma~\cite{argenziano2003dermoscopy}. 

\subsubsection{Dots and Globules} Dots are small structures of pigmented areas clustered in any distribution on or around the lesion. Dots clustered in center regions or on the network lines are referred to as regular, otherwise irregular. Globules are round, oval or polygonal structures larger than dots that can have high variability in colour, size and shape along with asymmetric distribution (Irregular Globules) or minimal variability along with symmetric distribution (Regular Globules). Regular Dots and Globules are indicators for benign melanocytic lesions and irregular Dots and Globules indicate melanoma \cite{argenziano2003dermoscopy}.

\subsubsection{Blue-Whitish Veils} Blue-Whitish Veils describe an irregularly shaped, structureless blotch on the lesion area that is characterized by a blue hue with an overlying whitish ground-glass haze. In~\cite{menzies1996sensitivity} it is rated as the most useful single diagnostic indicator for melanoma.

\subsubsection{Asymmetry} Asymmetry is the most important factor in malignant melanoma identification using ABCD rule~\cite{friedman1985early}. In our work, asymmetry refers to an asymmetrical lesion contour as well as asymmetrical distributions of structures and colours within a lesion~\cite{mendoncca2013ph}. The asymmetry concept is further divided into symmetric and asymmetric in one or two axes.

\subsubsection{Colour} This concept refers to colour present within the lesion area. As the appearance of single colours is not yet indicative of any diagnosis,  a combined concept of \textit{three or more colours} is used in the analysis. The presence of three or more colours increases the probability of melanoma drastically~\cite{argenziano2003dermoscopy}.
\bigbreak
The intricate explanations of concepts given above along with the concepts' innate variability offer much room for interpretation, implying the complexity of the problem itself. This is evident by the fact that even doctors tend to have notable disagreements when it comes to diagnosis, localization or identification of concept~\cite{fortina2012s, corona1996interobserver}.

\section{Materials \& Method}
\label{sec:MnM}

\subsection{Model}
\label{subsec:Model}
The model used in this work as the basis for our exploration and experimentation is developed by the University of Campinas in Brazil. Their RECOD Lab (REasoning for COmplex Data) made their submission~\cite{menegola2017recod} to the IEEE International Symposium on Biomedical Imaging (ISBI) 2017 challenge and is publicly available on github\footnote{https://github.com/learningtitans/isbi2017-part3}. By applying a transfer learning approach combined with extensive ensembling using an SVM meta-layer on top of seven base models trained on different data subsets, they achieved best Area Under Receiver Operating Characteristic (ROC) Curve (AUC) for Melanoma (MEL) classification (87.4\%), 3rd best AUC for Seborrheic Keratosis (SK) classification (94.3\%), and 3rd best combined/mean AUC (90.8\%) in part 3 of 2017 challenge. In this work, we intentionally refrained from training our own skin lesion classification model as our primary objective was explainability of these models instead of their classification performance. Thus, for our experimentation we only focused on a single module from RECOD's well-trained architecture. We used one of the base models\footnote{\textit{checkpoint.rc25 of RECOD model}} with Inception v4~\cite{szegedy2017inception} architecture, subsequently referred to as \textit{model} or \textit{network}. This base model was trained on RECOD's "deploy" set of 9,640 images using per-image normalization. 
\begin{figure*}[b!]
    \centering
    \includegraphics[width = 0.80\textwidth]{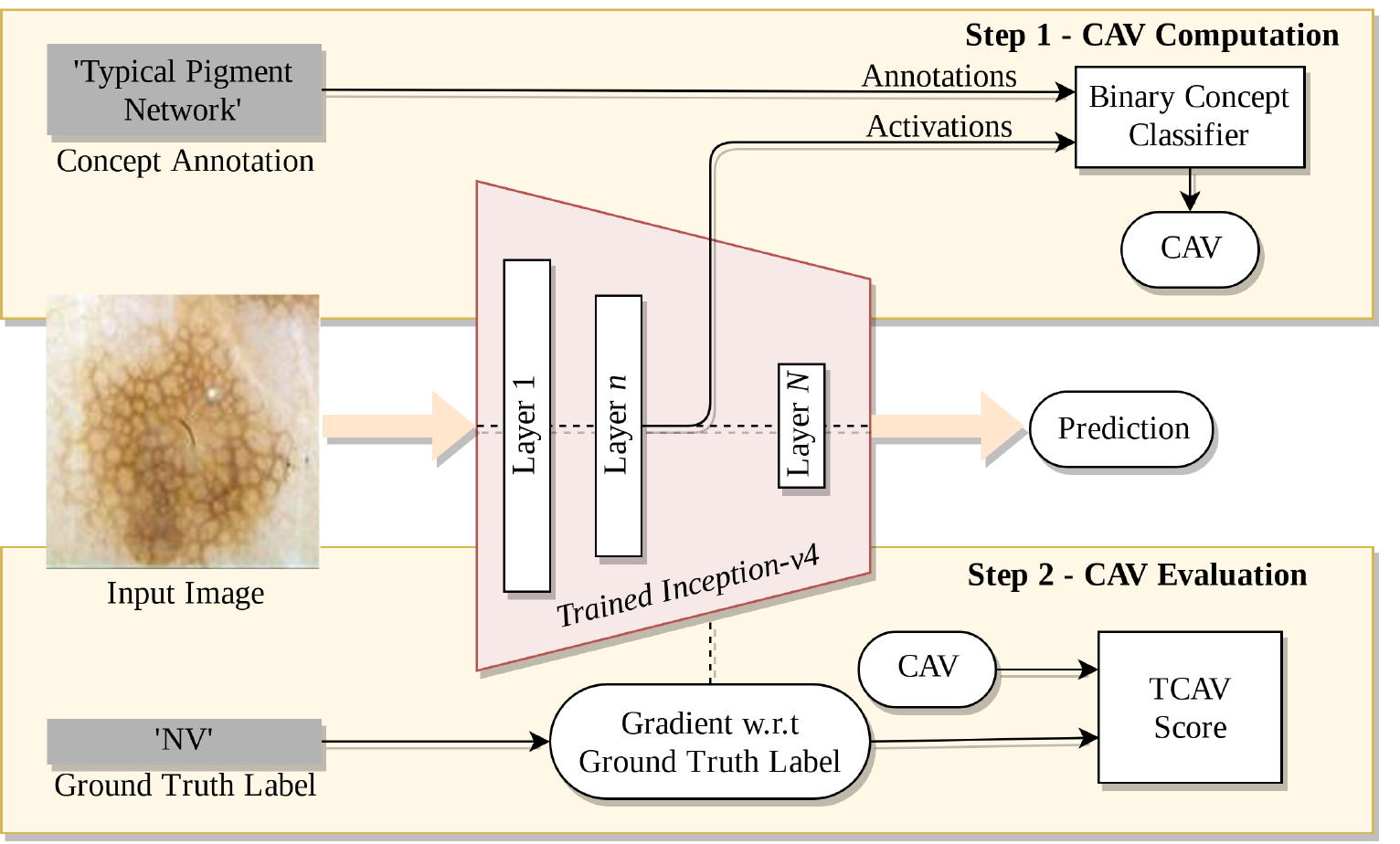}
    \caption{Overview of training concept classifiers and calculating CAV and TCAV scores.}
    \label{fig:architecture}
\end{figure*}

\subsection{Datasets}
\label{subsec:datasets}
The datasets used for concept training are PH\textsuperscript{2} dataset~\cite{mendoncca2013ph} and Seven-Point Checklist Dermatology dataset abbreviated as derm7pt~\cite{Kawahara2018-7pt}.

\begin{table}[b!]
\caption{Distribution of image samples into different concept classes in PH\textsuperscript{2} and derm7pt datasets. Note that PH\textsuperscript{2} dataset does not distinguish between regular and irregular streaks.}
\label{table:dataset} 
\centering
 \resizebox{\columnwidth}{!}{ 
\begin{tabular}{@{}cllccc@{}}
\toprule
\multicolumn{1}{c}{\textbf{Concepts}} & \multicolumn{2}{c}{\textbf{Presentation}} & \textbf{Abbreviation} & \multicolumn{1}{c}{\textbf{PH\textsuperscript{2}\cite{mendoncca2013ph}}} & \multicolumn{1}{c}{\textbf{derm7pt~\cite{Kawahara2018-7pt}}} \\ \midrule
\multirow{3}{*}{\textbf{\begin{tabular}[c]{@{}c@{}}Pigment\\ Network\end{tabular}}} & \multicolumn{2}{c}{} & PN & N/A & 551 \\
 & \multicolumn{2}{c}{Typical} & PN\_T & 84 & 335 \\
 & \multicolumn{2}{c}{Atypical} & PN\_AT & 116 & 216 \\ \midrule
\multirow{3}{*}{\textbf{Streaks}} & & & ST & 30 & 333 \\
 & \multicolumn{2}{c}{Regular} & ST\_R & N/A & 96 \\ 
 & \multicolumn{2}{c}{Irregular} & ST\_IR & N/A & 237 \\ \midrule
\multirow{2}{*}{\textbf{\begin{tabular}[c]{@{}c@{}}Regression\\ Structures\end{tabular}}} & \multicolumn{2}{l}{\multirow{2}{*}{}} & \multirow{2}{*}{RS} & \multirow{2}{*}{25} & \multirow{2}{*}{233} \\
 \\ \midrule
\multirow{3}{*}{\textbf{\begin{tabular}[c]{@{}c@{}}Dots \&\\ Globules\end{tabular}}} & \multicolumn{2}{c}{} & DG & 113 & 690 \\
 & \multicolumn{2}{c}{Regular} & DG\_R & 54 & 300 \\
 & \multicolumn{2}{c}{Irregular} & DR\_IR & 59 & 390 \\ \midrule
\multirow{2}{*}{\textbf{\begin{tabular}[c]{@{}c@{}}Blue-Whitish\\ Veils\end{tabular}}} & \multicolumn{2}{c}{\multirow{2}{*}{}} & \multirow{2}{*}{BWV} & \multirow{2}{*}{36} & \multirow{2}{*}{182} \\
 \\ \midrule
\multirow{3}{*}{\textbf{Asymmetry}} & \multicolumn{2}{c}{} & Sym & 117 & N/A \\
 & \multicolumn{2}{c}{1-Axis} & Asym\_1 & 31 & N/A \\
 & \multicolumn{2}{c}{2-Axis} & Asym\_2 & 52 & N/A \\ \midrule
\textbf{Colours} & \multicolumn{2}{c}{3 or more} & C\_3 & 39 & N/A \\ 
 \\ \midrule
\multicolumn{4}{c}{\textbf{Total Samples}} & 200 & 823 \\ \bottomrule
\end{tabular}%
}
\end{table}

The PH\textsuperscript{2} dataset consists of 200 dermoscopic images of melanocytic lesions, including 80 common naevi, 80 atypical naevi, and 40 melanomas. Along with the images, colour and lesion segmentation masks are provided as well as extensive well-curated annotations as seen in Table \ref{table:dataset}. The derm7pt dataset consists of 1,011 clinical and dermoscopic images. Each sample is assigned to either a miscellaneous class or one of 4 diagnosis classes. Two of these diagnosis classes i.e. Melanoma and Naevi (NV) are further divided into 13 sub-classes. From this dataset, only MEL and NV samples have been considered, resulting in 823 images. SK samples have been discounted due to their low count of only 45 samples. Table~\ref{table:dataset} provides an overview of number of samples for each concept class. 

For evaluation purposes, the original ISBI 2017 challenge dataset~\cite{codella2018skin} is used. The train set of ISBI 2017 challenge contains 1372 samples of NV, 374 samples of MEL and 254 samples of SK whereas test set contains 393 images of NV, 117 images of MEL and 90 images of SK.

In order to verify statistical significance of our results, we trained random CAVs to compare against our concept CAVs. For this purpose, random concept labels are assigned to a subset of the ISIC archive\footnote{https://isic-archive.com/ retrieved in November 2019} images, excluding MEL and NV classes, resulting in 2,870 samples. The idea behind leaving out those two classes is that remaining samples hardly contain concepts similar to the ones used for concept training.

\subsection{Experimental Setup}
As previously described, all experiments have been conducted on one of the Inception v4 base models from~\cite{menegola2017recod}. For each concept, binary classifiers are trained on network's activations to find the concepts' directions in the embedding space. The training and evaluation scheme is depicted in Fig.~\ref{fig:architecture}. First, activations are extracted from \textit{mixed\_6h} layer of the Inception v4 model using PH\textsuperscript{2} and derm7pt datasets. A clustering-based under-sampling technique as well as stratified splitting is applied to ensure evenly balanced train and validation sets for each binary concept training. 
Second, TCAV score is used to evaluate a concept's importance to a specific target class. To account for differences in pre-processing and classifier initialization, each classifier training is repeated  20 times on a randomly sampled dataset split, resulting in different CAVs and different TCAV scores. 

To check statistical significance of learned concepts, we trained additional 50 random CAVs per layer. The random datasets are produced by repeatedly sampling 1,000 random images from ISIC archive subset described in section~\ref{subsec:datasets} and assigning random binary labels. The distribution of random concept TCAV scores and actual concept TCAV scores is then compared by conducting a two-sided \emph{t}-test with $\alpha=0.05$ to assure significance of the found CAVs. In the results section, statistical insignificance is represented by red asterisks on top of the plotted bars.

\section{Results and Analysis}
\begin{figure*}[h!]
\centering
  \begin{subfigure}[h!]{0.49\textwidth}
  \centering
    \includegraphics[height=8.0cm]{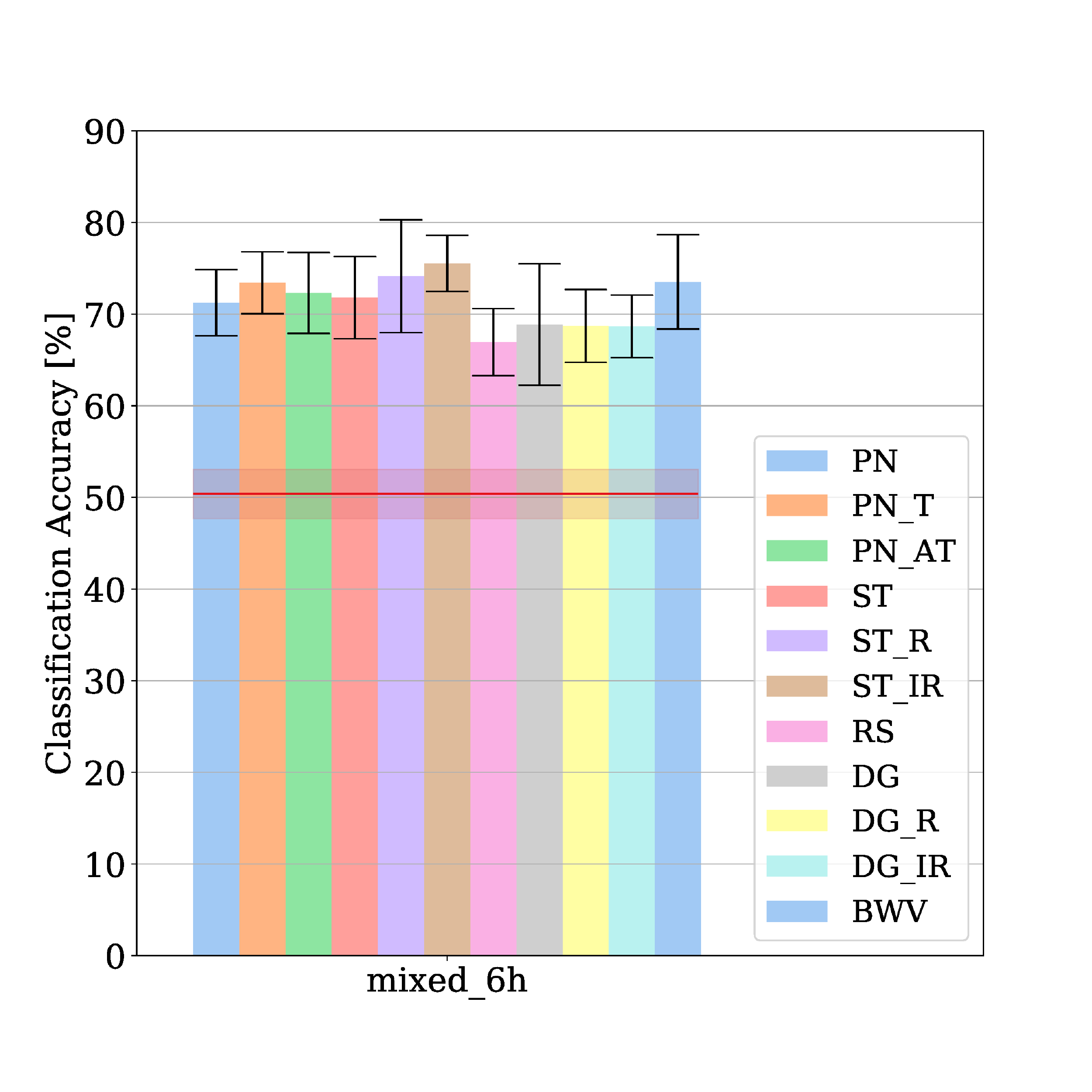}
    \caption{Derm7pt dataset}
    \label{fig:full_strat_derm7pt_acc}
  \end{subfigure}
  \begin{subfigure}[h!]{0.49\textwidth}
  \centering
    \includegraphics[height=8.0cm]{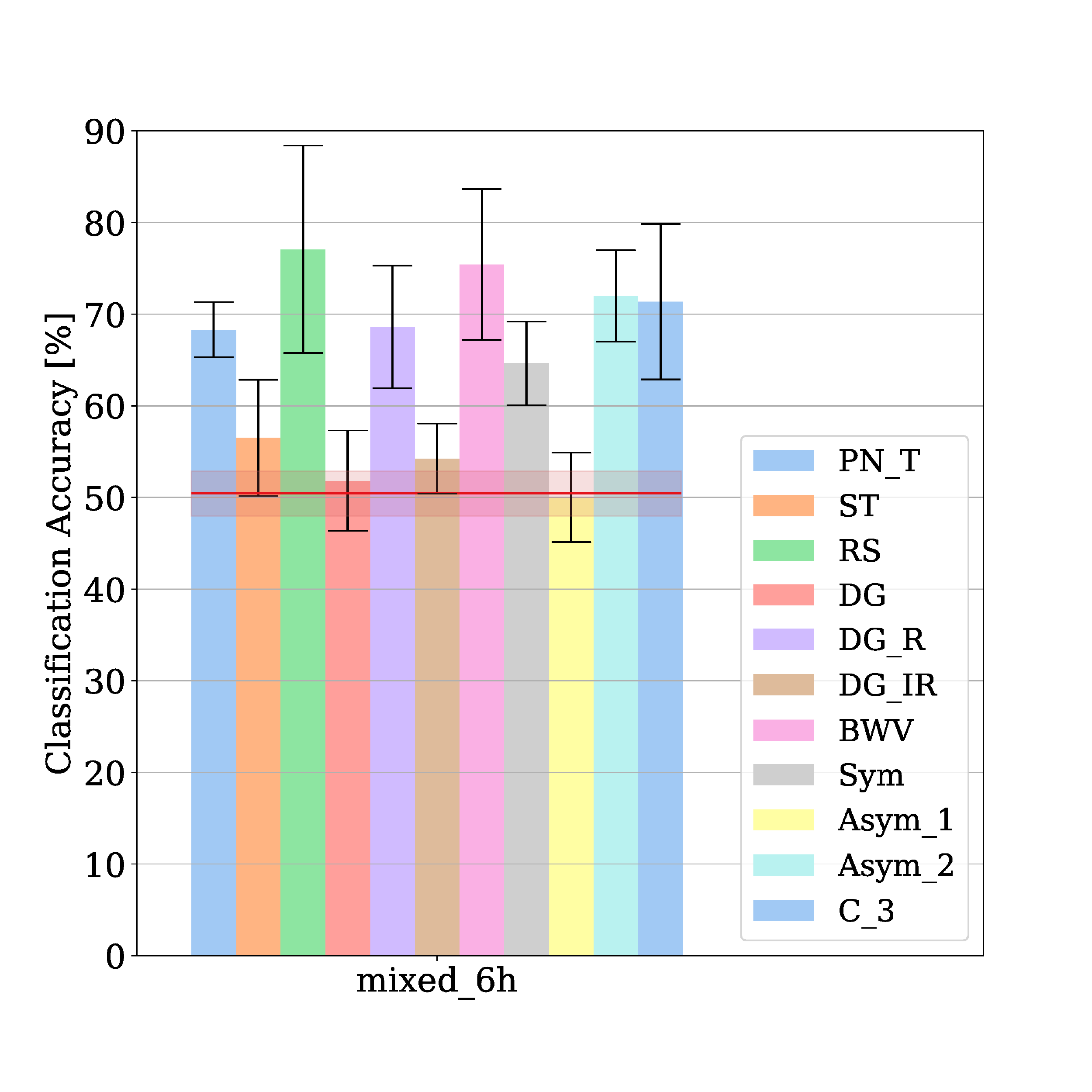}
    \caption{PH\textsuperscript{2} dataset}
    \label{fig:full_strat_ph2_acc}
  \end{subfigure}
  \caption{Validation accuracies of all concept classifiers trained and tested individually on derm7pt and PH\textsuperscript{2}. Random baseline is denoted by horizontal red line along with light red area marking standard deviation. Insignificant classifiers are marked with a red asterisk.}
  \label{fig:accuracies_validation_accs}
\end{figure*}

\begin{figure*}[b!]
    \centering
    \includegraphics[width = \textwidth]{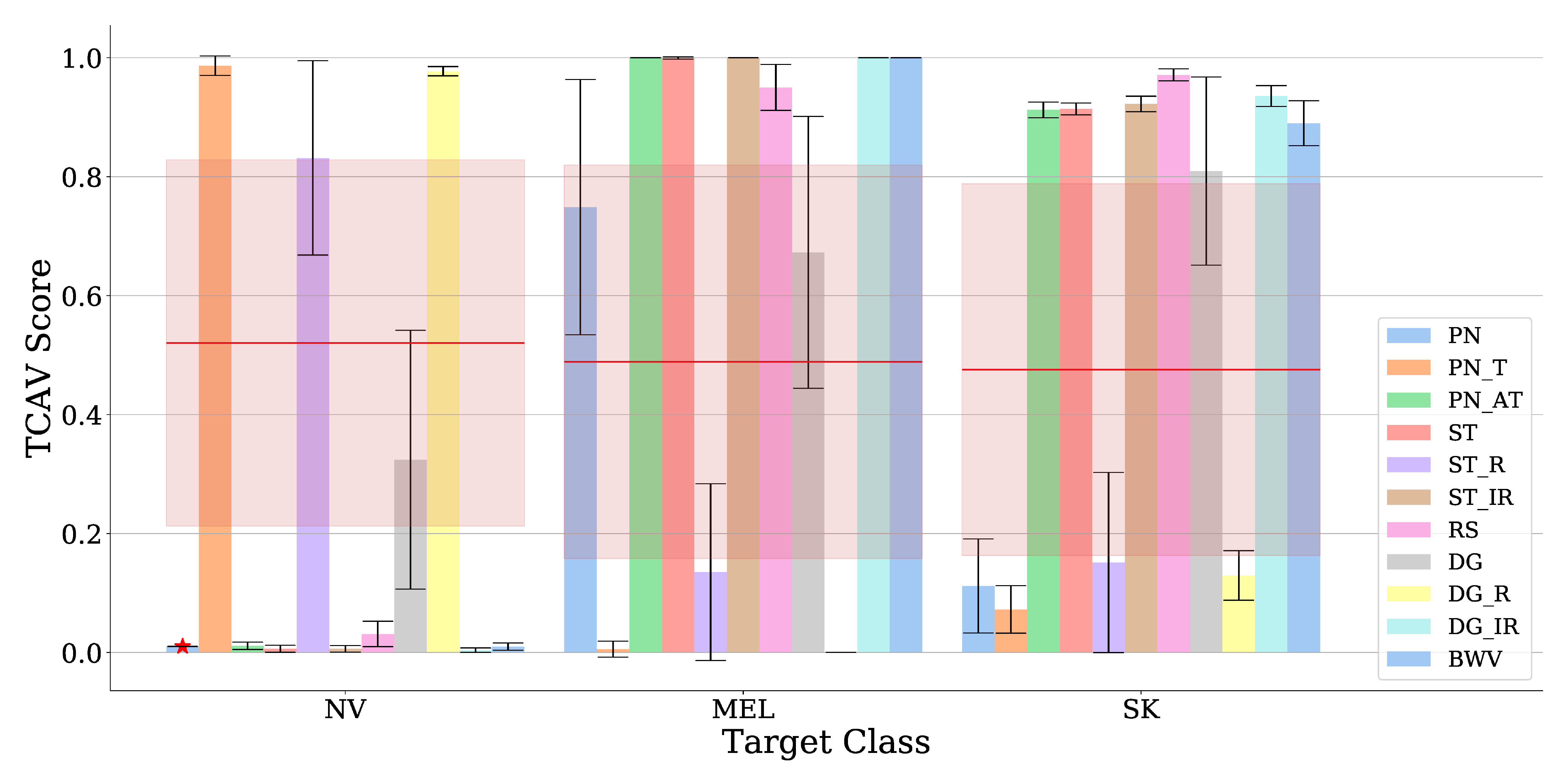}
    \caption{TCAV scores of each concept for derm7pt with respect to each target class on \textit{miexed\_6h} layer of RECOD model.}
    \label{fig:scores_derm7pt}
\end{figure*}

\begin{figure*}[t!]
    \centering
    \includegraphics[width = \textwidth]{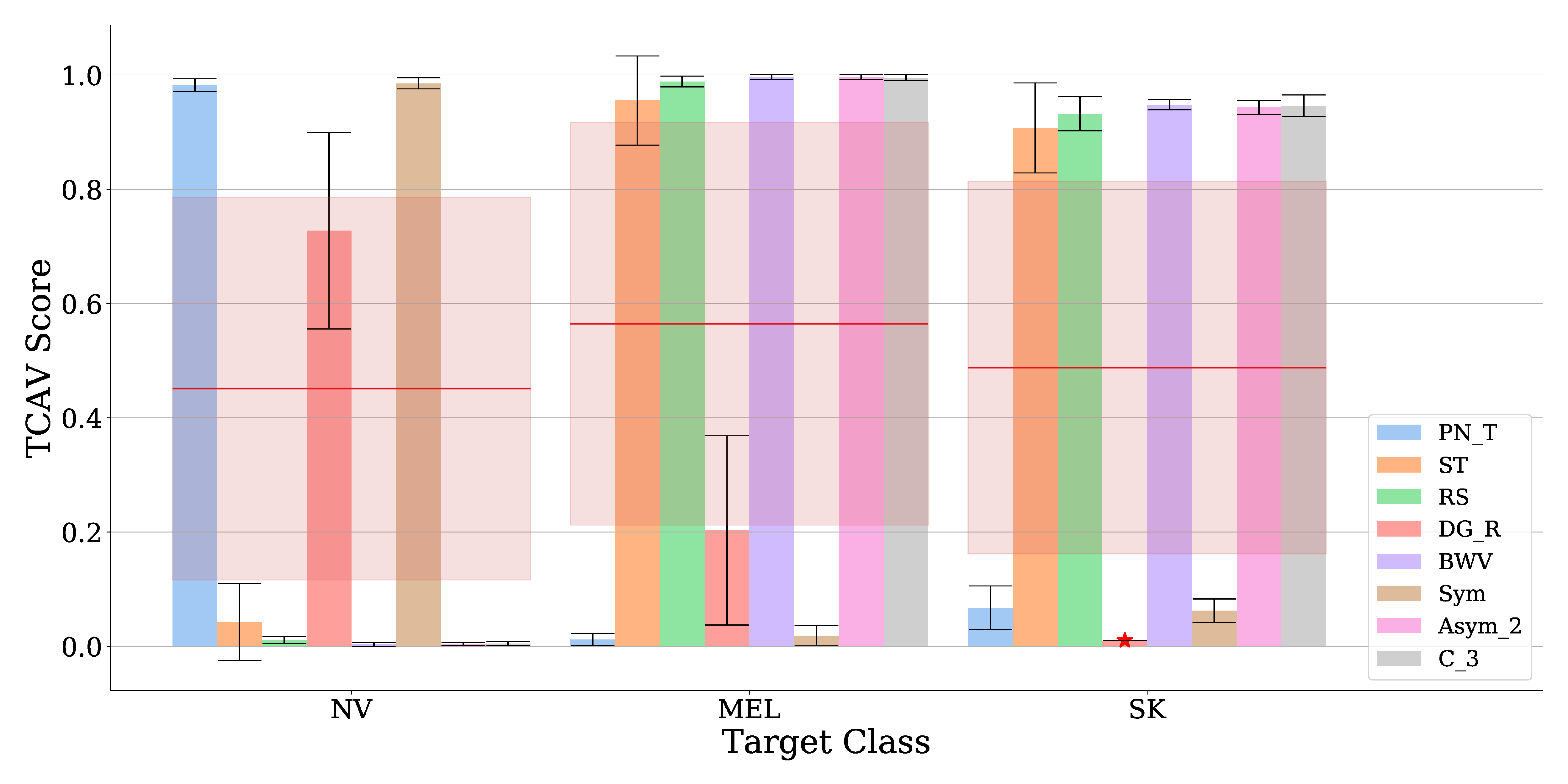}
    \caption{TCAV scores of each concept for PH\textsuperscript{2} with respect to each target class on \textit{mixed\_6h} layer of RECOD model.}
    \label{fig:scores_ph2}
\end{figure*}
The lack of quantifiability in most applicable explanation methods does not allow for proper comparison with previous approaches. Hence, we will focus on quantitative evaluation of concept classifier's accuracies and TCAV scores as well as a qualitative analysis of the resulting CAVs.

Fig.~\ref{fig:full_strat_derm7pt_acc} shows all mean validation accuracies achieved by individual binary concept classifiers along with their standard deviation trained on derm7pt embeddings from  \textit{mixed\_6h} layer. Mean baseline results from training on 50 random concept subsets are depicted by horizontal red line along with light red surrounding area marking standard deviation. It is evident from the figure that all concept classifiers achieved significantly higher validation accuracies than random baseline. The overall accuracies achieved might not seem very high, however it has to be mentioned  here that computation of CAVs requires the use of linear classifiers to calculate normal vector to decision hyperplane. The results are clear evidence that network's latent space is structured in a way that allows activation's separation with respect to similar concepts.

Fig.~\ref{fig:full_strat_ph2_acc} shows the classifiers' validation accuracy trained on PH\textsuperscript{2} dataset embeddings from \textit{mixed\_6h} layer. It is notable that many concepts achieved relatively mediocre accuracies near the random baseline. This can be explained by very small number of positive concept samples available in PH\textsuperscript{2} dataset. 

The TCAV score quantifies positive or negative influence of a given concept towards a specific target class. Values above 0.5 indicate positive influence of the concept to the prediction and lower values indicate negative influence. Figures~\ref{fig:scores_derm7pt} and~\ref{fig:scores_ph2} show the TCAV scores achieved by evaluating 20 CAVs per concept on the \textit{mixed\_6h} layer separately trained on both datasets. Average baseline scores of all 50 random concepts are again depicted by red horizontal lines along with their standard deviation in light red. Statistically insignificant results are marked by red asterisks. 

Results for NV and MEL classes for concepts trained using derm7pt look very much as expected. Although, the score for \textit{PN} turned out to be insignificant in one experiment, features indicating benign melanocytic lesions like \textit{PN\_T}, \textit{ST\_R} and \textit{DG\_R} all contributed positively towards NV class. On the other hand, strong signs for malignant melanoma like \textit{PN\_AT}, \textit{ST\_IR}, \textit{RS}, \textit{DG\_IR} and \textit{BWV} show strong negative influence. Also, it is notable that the presence of Streaks in general (\textit{ST}) has a stronger negative influence as compared to the presence of particularly regular Streaks (\textit{ST\_R}). Results for MEL class show the exact opposite behaviour, which is perfectly aligned with the descriptions in medical literature. It is again notable that the presence of Dots and Globules (\textit{DG}) and the presence of Streaks (\textit{ST}) show higher positive impact on MEL class as compared to their regular variants, for example, regular Streaks (\textit{ST\_R}). Results for SK class show similar concept influence as for MEL, except for Pigment Networks (\textit{PN}) exhibiting negative influence. In~\cite{braun2002dermoscopy} the appearance of network-like structures in Seborrheic Keratosis has been confirmed. The model might have encoded those structures in the atypical Pigment Network (\textit{PN\_AT}) concept, as their appearance slightly differs from the classical Pigment Networks definition. In the same study, evidence for Dots and blue-gray areas in SK lesions have been found as well.

Fig.~\ref{fig:scores_ph2} shows resulting TCAV scores for CAVs trained on PH\textsuperscript{2} dataset. All concepts achieving less than 55\% validation accuracy have not been considered. Again, TCAV scores for NV and MEL show expected behaviour. Only typical Pigment Networks (\textit{PN\_T}), regular Dots and Globules (\textit{DG\_R}) and Symmetry (\textit{Sym}) contribute positively towards Naevi class. For melanoma, the exact opposite holds again which can be confirmed by the concept descriptions in Section~\ref{subsec:concepts}. Additionally, from the results it appears that asymmetric lesions (\textit{Asym\_2}) and lesions containing more than three colours (\textit{C\_3}) tend to be classified as melanoma. For SK we can again observe low influence of typical Pigment Networks (\textit{PN\_T}) as well as high influence for all other concepts including asymmetry (\textit{Asym\_2}) and colour diversity (\textit{C\_3}).

\begin{figure*}[h!]
    \centering
    \includegraphics[width = 0.87\textwidth]{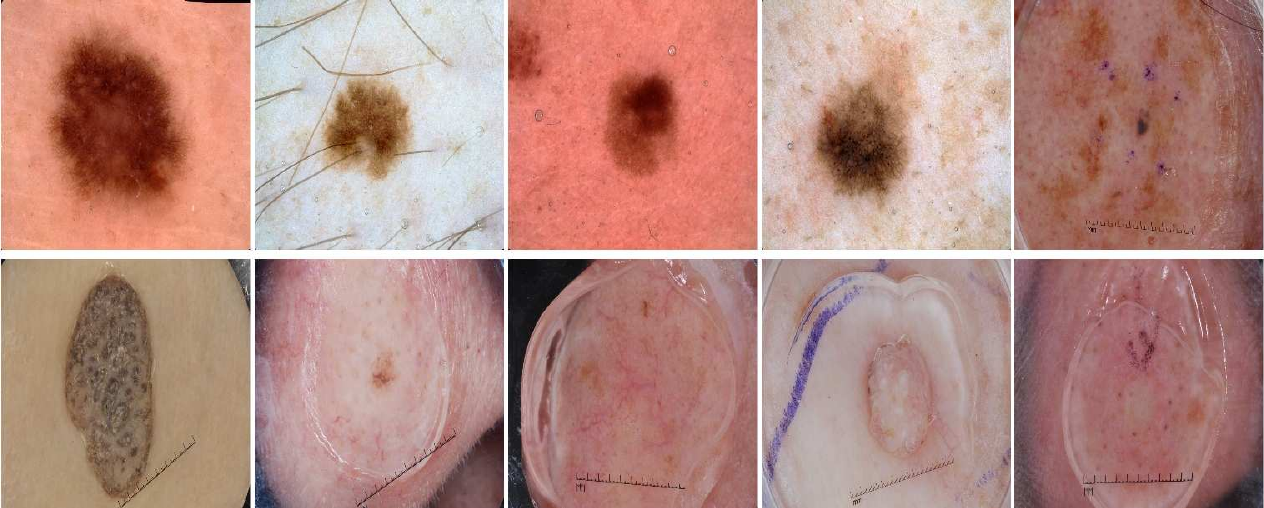}
    \caption{First row shows the test images that are the most similar to Typical Pigment Network (\textit{PN\_T}) concept according to euclidean distance in CAV direction. The second row shows the least similar images to PN\_T.}
    \label{fig:sprite_PNT}
\end{figure*}

\begin{figure*}[h!]
    \centering
    \includegraphics[width = 0.87\textwidth]{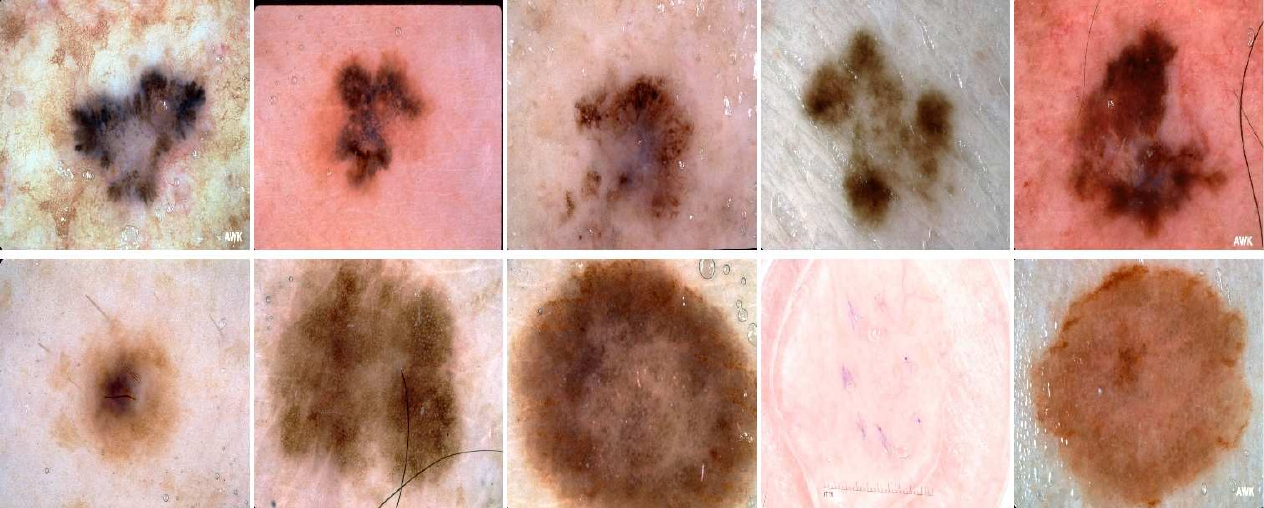}
    \caption{First row shows the test images that are the most similar to Irregular Streaks (\textit{ST\_IR}) concept according to euclidean distance in CAV direction. The second row shows the least similar images to ST\_IR.}
    \label{fig:sprite_DGIR}
\end{figure*}

\begin{figure*}[h!]
    \centering
    \includegraphics[width = 0.87\textwidth]{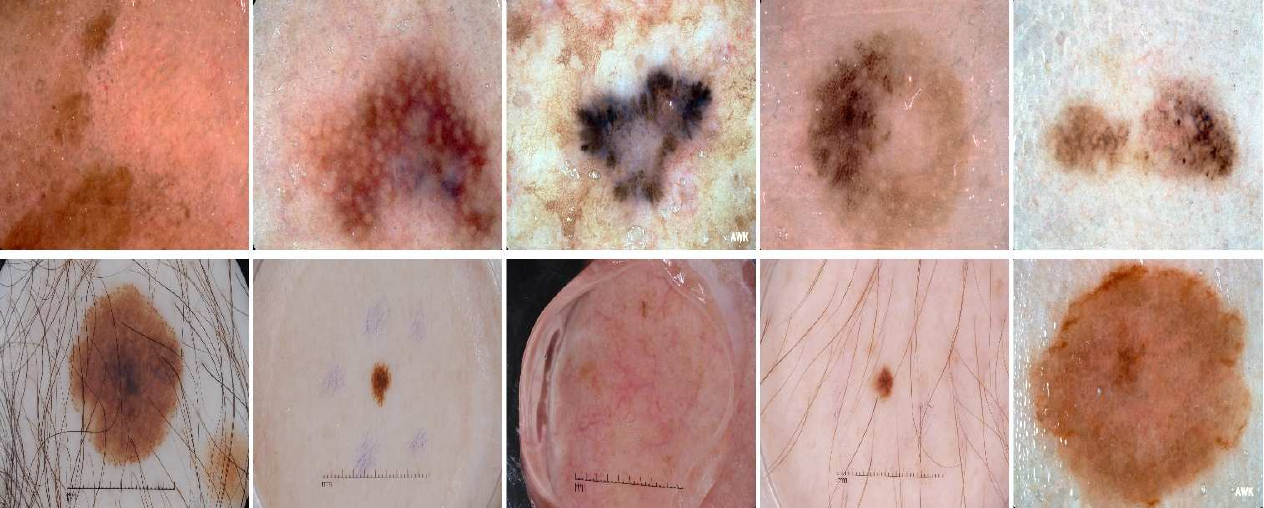}
    \caption{First row shows the test images that are the most similar to Regression Structure (\textit{RS}) concept according to euclidean distance in CAV direction. The second row shows the least similar images to RS.}
    \label{fig:sprite_RS}
\end{figure*}

To further validate that the model has comprehensively learnt these disease-related concepts instead of learning some random concepts, we had our model sort all the test images with respect to degree of visibility of a certain concept in each image. In other words, the model ordered all 300 test images starting from those that presented very obvious existence of a concept and ending with those which had least evidence of that concept. This ordering is performed based on euclidean distance in a CAV's direction. Fig.~\ref{fig:sprite_PNT} through Fig.~\ref{fig:sprite_RS} show first five and last five images from the sorted test set with respect to different concepts. The first row of each figure shows positive examples, where the concept is most clearly visible, and the second row shows negative examples, where the concept is virtually absent. It is evident from these figures that the proposed method for explaining skin disease classifiers does not only provide justification of classifier's decision on global dataset scale but also sensibly identifies reasons for per-image predictions.


\section{Conclusion}
Concept-based methods for network explanation offer great potential especially for complex classification tasks in critical application areas like MIA. This work strives to leverage these methods to verify the ability of DNNs to learn and utilize human understandable concepts for skin lesion classification. Our results show strong correlation between DNN's learnt representation of various concepts and those routinely used by dermatologists. This work corroborates that deep learning based CAD systems are able to learn and utilize similar disease-related concepts for prediction as used by dermatologists. We hope that physicians would be able to confer higher confidence to such CAD systems that are able to justify their prediction by listing the concepts which influenced positively or negatively towards a certain output. It has also been shown that Testing with CAVs (TCAV) is applicable using complete identically distributed images instead of general concept patches. However, this work can further be improved by using more granular labelling of diseases indicative concepts to get deeper insight into model's classification processes as well as further validation of its decisions. 

Due to the complexity of the problem, possibly subjective annotations of training set by various expert and small number of concept training samples, not all human-defined disease-related concepts were thoroughly analysed. Standardizing the annotation according to one \emph{school of thought} in dermatology community, for example following \cite{kittler2016standardization}, can decrease inter-observer disagreement but it would require an enormous amount of time and effort by dermatology experts. Nevertheless, the obtained results are remarkably aligned with common diagnostic criteria. In order to allow for a more comprehensive interpretation of the TCAV scores for this specific task, it would be desirable to curate a high-quality dataset with reliable fine-grained labels of concepts that are known to be highly indicative of specific diagnoses.

Supervised concept learning is highly dependent on high quality and precisely annotated human concept examples. Therefore, more focus should be placed on generating clean datasets of high-quality concept annotations that can be used for explaining models in medical imaging applications to speed up their deployments in clinical routines. As supervised concept classification from network activations has already been proven to be effective, the following logical and highly desirable extension of unsupervised concept discovery should be considered. Not only would this be another improvement towards simplifying the interpretability of networks by eliminating the necessity for laborious expert annotations but it could also allow insights in a network's own concepts, potentially revealing new knowledge for domain experts or unexpected biases in the network.


\bibliographystyle{IEEEtran}
\bibliography{references.bib}

\end{document}